\documentclass[10pt,twocolumn,letterpaper]{article}

\usepackage{btas}
\usepackage{times}
\usepackage{epsfig}
\usepackage{graphicx}
\usepackage{amsmath}
\usepackage{amssymb}
\usepackage{multirow}
\usepackage{tabularx}
\usepackage{subfigure}
\graphicspath{{fig/}}

\btasfinalcopy % *** Uncomment this line for the final submission

\ifbtasfinal\fi
\begin{document}

%%%%%%%%% TITLE
\title{FingerNet: An Unified Deep Network for Fingerprint Minutiae Extraction}

\author{Yao Tang \quad Fei Gao \quad Jufu Feng* \quad Yuhang Liu\\
Key Laboratory of Machine Perception (MOE), School of EECS, Peking University\\
{\tt\small \{tangyao@, goofy@, fjf@cis., liuyuhang@\}pku.edu.cn}
}

\maketitle
\thispagestyle{empty}

%%%%%%%%% ABSTRACT
\begin{abstract}
Minutiae extraction is of critical importance in automated fingerprint recognition. Previous works on rolled/slap fingerprints failed on latent fingerprints due to noisy ridge patterns and complex background noises.
In this paper, we propose a new way to design deep convolutional network combining domain knowledge and the representation ability of deep learning.
In terms of orientation estimation, segmentation, enhancement and minutiae extraction, several typical traditional methods performed well on rolled/slap fingerprints are transformed into convolutional manners and integrated as an unified plain network. We demonstrate that this pipeline is equivalent to a shallow network with fixed weights. The network is then expanded to enhance its representation ability and the weights are released to learn complex background variance from data, while preserving end-to-end differentiability.
Experimental results on NIST SD27 latent database and FVC 2004 slap database demonstrate that the proposed algorithm outperforms the state-of-the-art minutiae extraction algorithms. Code is made publicly available at: https://github.com/felixTY/FingerNet.
\end{abstract}

%%%%%%%%% BODY TEXT
\section{Introduction}

Minutiae are the premier features in most fingerprint matching systems~\cite{jain1997line}. Extracting minutiae from rolled/slap fingerprints has been studied for many years, and acquires reliable results.
However the accuracy degrades significantly in latent fingerprints because of
fuzzy ridges and complex background noises.
Latent fingerprints are obtained directly from crime scenes and their minutiae are manually marked by experts. It is of great significance to obtain valuable information from the fingerprints left on the scene.
Extensive research has been undertaken on latent fingerprints in various fields.
Table.~\ref{table:related} summarizes some studies on latent fingerprints in recent years.

%Therefore, reliable and efficient minutiae extraction is an extremely important task.
%In the past, minutiae on latent fingerprints are marked by experts, and then matched to find the suspect. So an automated minutiae extraction algorithm can greatly reduce the labor cost, improve the work efficiency and obtain more reliable results.

%Latent fingerprints are obtained directly from the crime scene. It is of great significance to obtain valuable information from the fingerprints left on the scene. Minutiae are the premier features in most fingerprint matching systems~\cite{jain1997line}.
%%Therefore, reliable and efficient minutiae extraction is an extremely important task.
%In the past, minutiae on latent fingerprints are marked by experts, and then matched to find the suspect. The automated minutiae extraction algorithm can greatly reduce the labor cost, improve the work efficiency and obtain more reliable results.
%
%Extracting minutiae from rolled/slap fingerprints has been studied for many years, and acquired reliable results. It is also a critical step in an automated fingerprint identification system. However the accuracy degrades significantly in latent fingerprints because of
%fuzzy ridges and complex background noises.
%Extensive research has been undertaken on latent fingerprints in various fields.
%Table.~\ref{table:related} summarizes some studies on latent fingerprints in recent years.

\setlength{\tabcolsep}{6pt}
\renewcommand\arraystretch{1.1}
\begin{table*}[t]
\begin{center}
\label{table:related}
%\scriptsize
\footnotesize
\newcolumntype{Y}{>{\centering\arraybackslash}X}
\begin{tabularx}{0.9\textwidth}{|c|Y|p{6cm}<{\centering}|Y|}
\hline
Modules & Study & Approach & Database \\
\hline
\hline
\multirow{2}*{Segmentation} & Ruangsakul et al.~\cite{ruangsakul2015latent} & Rearranged fourier subbands & NIST SD27 \\ \cline{2-4}
                            & Choi et al.~\cite{choi2012automatic} & Combined local ridge frequency and orientation & NIST SD27 and WVU \\
\hline
Orientation & Cao et al.~\cite{cao2015latent} & convolutional neural network & NIST SD27\\
\hline
Enhancement & Cao et al.~\cite{cao2014segmentation} & ridge structure dictionary & NIST SD27 and WVU \\
\hline
\multirow{2}*{Extraction} & Sankaran et al.~\cite{sankaran2014latent} & Stacked denoising sparse autoencoders & NIST SD27 \\ \cline{2-4}
                          & Tang et al.~\cite{tang2016latent} & Fully convolutional network & NIST SD27\\
\hline
\multirow{2}*{Matching} & Jain et al.~\cite{jain2011latent} & Local and global matching with extended features & NIST SD27 \\ \cline{2-4}
                        & Paulino et al.~\cite{paulino2013latent} & Descriptor-based hough transform & NIST SD27 and WVU \\
\hline

\end{tabularx}
\end{center}
\caption{
	Recent studies and their main approaches on latent fingerprints.
}
\end{table*}
\setlength{\tabcolsep}{1.4pt}

\begin{figure}[t]
\centering
\subfigure[latent fingerprint]
{\label{fig:1}\includegraphics[width=0.4\linewidth]{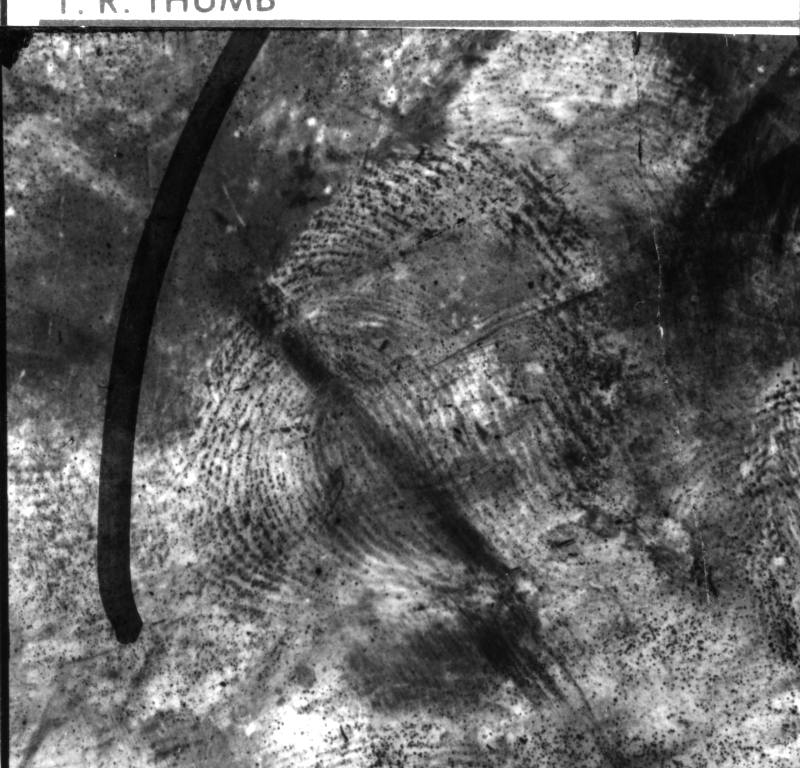}}
\subfigure[orientation field]{\label{fig:2}\includegraphics[width=0.4\linewidth]{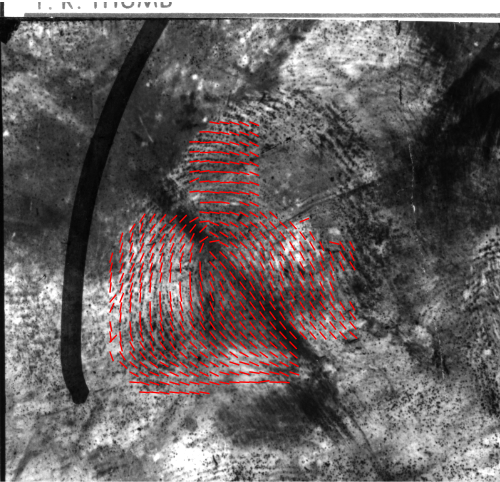}}
\subfigure[enhanced fingerprint]{\label{fig:3}\includegraphics[width=0.4\linewidth]{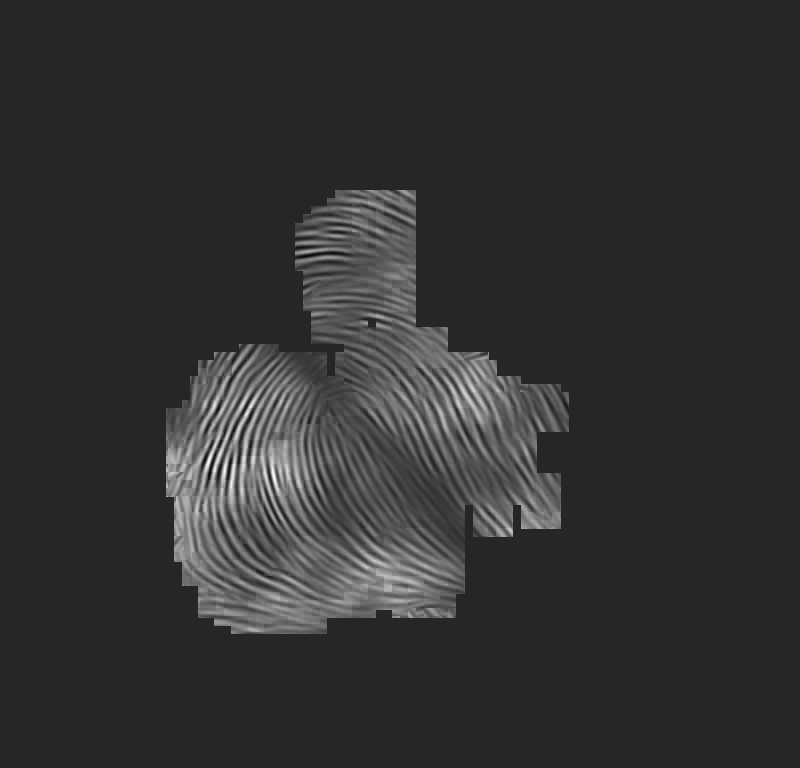}}
\subfigure[minutiae map]{\label{fig:4}\includegraphics[width=0.4\linewidth]{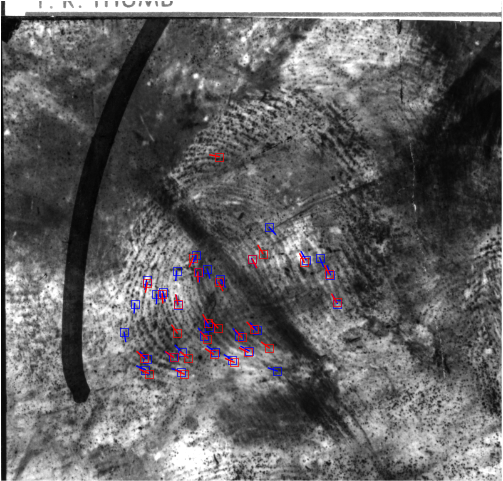}}
\caption{Sample results of our proposed FingerNet. (a) input latent fingerprint, (b)-(d) output orientation field, enhanced fingerprint and minutiae map. In (d), red squares denote extracted minutiae and blue squares denote manually marked minutiae.}
\label{fig:sample}
\end{figure}

So far, minutiae extraction methods can be divided into two categories. One is traditional method using handcrafted features designed by domain knowledge.
Ratha et al.~\cite{ratha1995adaptive} followed the simple idea of ridge extraction, thinning and minutia extraction.
%So algorithm based on ridge extraction and thinning does not work well in latent fingerprints.
Gao et al.~\cite{gao2010novel} extracted minutiae on Gabor phase, which means the fingerprints have been enhanced with Gabor filters to overcome the influence of creases and noises.
But in latent fingerprints, handcrafted features are difficult to adapt to complex background variance.
%However, Gabor filters are heavily dependent on the calculate of ridge orientation and frequency. Calculating orientation field on latent fingerprints is also a hard task as described in~\cite{cao2014segmentation,yang2014localized,cao2015latent}.
Another is deep learning method which learns features from data automatically.
Sankaran et al.~\cite{sankaran2014latent} used stacked denoising sparse autoencoders to learn features to classify minutiae and non-minutiae patches.
%The DSAE learns features from data rather than handcrafted features.
%They have gained a promising results on latent fingerprints.
%The direction of minutia is also ignored which is really important information.
Tang et al.~\cite{tang2016latent} regarded minutiae extraction as an object detection task. They extracted minutiae from a learned fully convolutional network.
However the domain knowledge in fingerprints is not considered in these methods, such as the basic hypothesis of 2D amplitude and frequency modulated (AM-FM) signal~\cite{larkin2007coherent}.
Transferring the network learned in natural images to fingerprints seemed to limit their performances.
%Traditional methods are designed by prior knowledge indicating fingerprints' inherent properties. It performed well on rolled fingerprints but failed on latent due to noisy ridge pattern and presence of background noise. Learning based methods learn the complex background variance from data. So it improves the performance on bad quality fingerprints.

%Deep convolutional networks have outperformed many traditional methods on computer vision areas, such as segmentation, localization and object detection~\cite{sermanet2013overfeat,long2015fully,ren2015faster,redmon2016you,liu2016ssd,kong2016hypernet,liu2016accurate}.
Our basic idea is to combine domain knowledge and deep learning representation ability.
Some researchers designed special structures with domain knowledge in specific areas, such as smoothing, denoising, inpainting and color interpolation.
Liu et al.~\cite{liu2016learning} transformed infinite impulse response filters into recurrent neural networks and learned the weights by a deep convolutional neural network. They achieved promising results through a simpler and faster network. Ren et al.~\cite{ren2015shepard} demonstrated that the translation variant interpolation can not be simply modeled by a single kernel due to the inherent spatially varying property, so they designed a Shepard interpolation layer as translation variant operations for inpainting.

In this paper, a new way is proposed to guide the network's structure design and weight initialization combing both traditional methods and deep convolutional networks.
%An universal deep convolutional network is designed to extract minutiae.
We demonstrate that the minutiae extraction pipeline consisting of orientation estimation, segmentation, Gabor enhancement and extraction is equivalent to a simple network with fixed weights, thus their representation ability is limited and they can't learn complex background noise from latent fingerprints.
%Typical traditional methods on orientation estimation, segmentation, Gabor enhancement and minutiae extraction are transformed into convolutional manner, and then integrated as a simple convolutional network with fixed weights.
Naturally, the simple network is then expanded with some convolutional layers to enhance its representation ability, and the weights are released to learn complex background from data.
%That means traditional methods are utilized as domain knowledge to guide the structure design and weights initialization.
The specially designed network for fingerprints is called FingerNet.
Benefiting from our design idea, the mechanism of FingerNet can be understood and typical fingerprint representations including orientation field, segmentation and enhancement can be acquired during minutiae extraction.

Considering the lack of training labels of orientation or segmentation, weak labels are generated based on the matching of latent fingerprints and corresponding rolled/slap fingerprints.
Fig.~\ref{fig:sample} shows a sample for orientation estimation, segmentation, enhancement and extraction on a latent fingerprint.
We also get promising performance on good quality fingerprints like FVC 2004 database~\cite{maio2004fvc2004}.

The key contributions of this paper are as follows:
\begin{enumerate}
    \item A new way to guide the deep network's structure design and weight initialization to combine domain knowledge and the representation ability of deep learning, while preserving end-to-end differentiability.
    \item A novel network for fingerprints called FingerNet is proposed. Typical fingerprint representations including orientation field, segmentation, enhancement and minutiae can be acquired from the unified network.
    \item Reliable minutiae have been extracted on both rolled/slap and latent fingerprints automatically without any fine tuning.
%    \item Typical traditional methods on orientation estimation, segmentation, Gabor enhancement and minutiae extraction are transformed into convolutional manner, and thus they can be embedded in popular network structures.
    \item One way to generate weak labels to latent fingerprints from the matched rolled/slap fingerprints, which helps to achieve modular training.
\end{enumerate}

\begin{figure*}[t]
 \centering
 \includegraphics[width=0.8\linewidth]{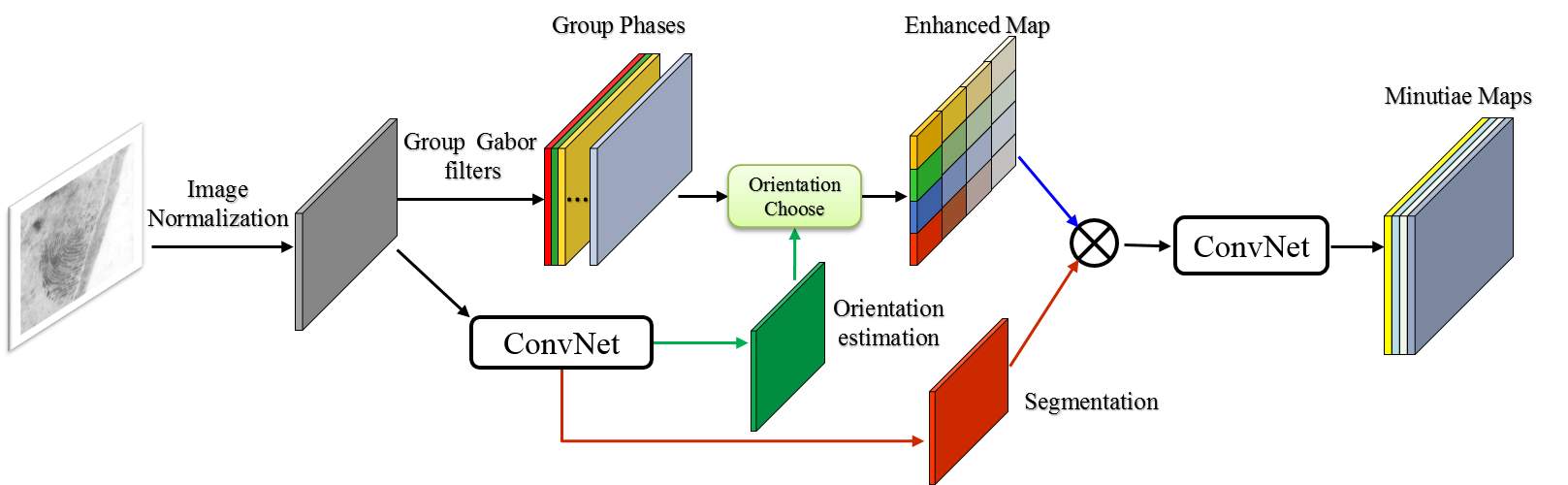}
 \caption{Our integrated minutiae extraction pipeline. Traditional methods consisting of orientation estimation, segmentation, Gabor enhancement and extraction are transformed into convolutional manners and integrated as a network.}
 \label{fig:pipeline}
\end{figure*}
%-------------------------------------------------------------------------
\section{Proposed FingerNet for Minutiae Extraction}

The basic idea is to build a specific network for fingerprints, which integrates the essence of traditional handcrafted methods (domain knowledge) and the representation ability of deep learning. We first transform some traditional methods to convolutional kernels and integrate them as a shallow network. This network is shallow and has fixed weights. The entire procedure integrating normalization, orientation estimation, segmentation, Gabor enhancement and minutiae extraction is visible in Fig.~\ref{fig:pipeline}. Next, we discuss how to expand the plain network to a complete trainable network for fingerprints.
We then describe weak label, loss definition and training procedure in detail.

\subsection{Traditional Methods to Equivalent ConvNets} \label{section:transform}

Traditional fingerprint minutiae extraction pipeline can be summarized as: normalization, orientation estimation, segmentation, enhancement and minutiae extraction.
%Since most of the digital image processing operations are convolutional operations, a majority of fingerprint methods can be equivalently transformed into convolutional networks (ConvNets).
Here, we transform several classical methods and construct a ConvNet, called plain FingerNet as an example. The plain FingerNet pipeline and connection relationships can be seen in Fig~\ref{fig:pipeline}.

It should be noted that all the operators in this article are pixel-wise operators and differentiable.

\subsubsection{Normalization} \label{section:fingerprint normalization}
One pixel-wise method~\cite{hong1998fingerprint} adjusts the intensity value of each pixel to a same scale as,
\begin{align}
    I'(x,y)&=\begin{cases}
    m_{0}+\sqrt{\frac{(I(x,y)-m)^2\cdot v_{0}}{v}},~I(x,y)>m \\
    m_{0}-\sqrt{\frac{(I(x,y)-m)^2\cdot v_{0}}{v}},~otherwise
    \end{cases}
\end{align}
where $I(x,y)$ is the intensity value at pixel $(x,y)$ in input image $I$, $m$ and $v$ are the image mean and variance and $m_0$ and $v_0$ are the desired mean and variance after the normalization.

This normalization operation can be regarded as a nonlinear nonparametric pixel-wise activation layer in plain FingerNet. Similar idea can be found in LRN layer~\cite{krizhevsky2012imagenet} and BatchNorm layer~\cite{ioffe2015batch}.

\subsubsection{Orientation Estimation}
By replacing gradient computation and sum of windowed value with convolutional operations, the gradient-based orientation estimation method~\cite{ratha1995adaptive} computing ridge orientation can be transformed as,
\begin{align}
\triangledown_x I&= I \ast S_x, ~\triangledown_y I=I \ast S_y, \notag \\
G_{xy} &= (\triangledown_x I \cdot \triangledown_y I) \ast J_w, \notag \\
G_{xx} &= (\triangledown_x I)^2 \ast J_w, \notag \\
G_{yy} &= (\triangledown_y I)^2 \ast J_w, \notag \\
\theta &= 90^{\circ}+\frac{1}{2}atan2(2 \cdot G_{xy},G_{xx}-G_{yy}), \label{eq:GGG}
\end{align}
where $\triangledown_x$ and $\triangledown_y$ are the $x-$ and $y-$ gradients computed through Sobel masks $S_{x}$ and $S_{y}$, $*$ indicates a convolutional operator, $J_w$ is an all-ones matrix with size of $w\times w$, $atan2(y,x)$ calculates the arc tangent of the two variables y and x with consideration of their quadrant and $\theta$ is the output orientation field .

It is actually a shallow ConvNet with 3 handcrafted kernels, a few merge layers and complex activation layers.

\subsubsection{Segmentation} \label{section:segmentation}
%Segmentation can be summarized as two steps: quality evaluation and threshold filter, corresponding to features extraction and bias choose in a ConvNet.
One learning-based segmentation method~\cite{bazen2001segmentation} trains a linear classifier based on handcrafted features like gradient coherence, local mean, and local variance.
%Similar to orientation estimation part of plain FingerNet, we first transform the calculation of handcrafted features into convolutional manner and concatenate them on channel dimension to form a feature map. And then transform the pixel-wise linear classifier to a $1 \times 1$ convolutional operation~\cite{long2015fully}.
This method can be computed as,
\begin{align}
Coh&=\frac{\sqrt{(G_{xx}-G_{yy})^2+4 \cdot G_{xy}^2}}{G_{xx}+G_{yy}},  \notag \\
Mean&=\frac{I \ast J_w}{w^2}, ~Var=\frac{(I-Mean)^2 \ast J_w}{w^2}, \notag \\
Seg &= \omega \ast [Coh, Mean, Var]+\beta,
\end{align}
where $w$ is the length of local window, $\omega$ and $\beta$ are the classifier's parameters and $[\cdot,\cdot,\cdot]$ indicates concatenation on channel dimension.

It is also a shallow ConvNet, and shares $G_{\cdot\cdot}$ with orientation estimation part as defined in Eq.~\ref{eq:GGG} .

\subsubsection{Enhancement} \label{section:gabor enhancement}
Gabor enhancement~\cite{bernard2002fingerprint} is widely used in fingerprint recognition systems because of its frequency-selection characteristic.
%Its complex form can be expressed as,
%\begin{equation}
%    g_{\omega,\theta}(x,y)=\frac{1}{\sqrt{\sigma_{\theta\bot}}}e^{-\frac{v^{2}}{2\sigma^{2}_{\theta\bot}}}\cdot
%    \frac{1}{\sqrt{\sigma_{\theta}}}e^{-\frac{u^{2}}{2\sigma^{2}_{\theta}}}\cdot e^{i\omega u},
%\end{equation}
%
%where
%\begin{equation}
%    u=x\cos\theta+y\sin\theta,
%\end{equation}
%\begin{equation}
%    v=-x\sin\theta+y\cos\theta,
%\end{equation}
%Here, $\theta$ and $\omega$ denote the orientation and frequency of the filter, respectively. $\sigma_{\theta}$ and $\sigma_{\theta\bot}$ are standard deviation along the direction of $\theta$ and its perpendicular direction.
The complex Gabor filter $g_{\omega,\theta}$ is generated from local ridge frequency $\omega$ and local ridge orientation $\theta$, then convolution operations are conducted on local fingerprint block. The enhanced complex block $E_D$ can be described as follows.

For each pixel $(x,y)$ in block $I_{D}$,
\begin{align}
    E_D(x,y)&=(I_{D}\ast g_{\omega,\theta})(x,y) \\
              &=A(x,y)\cdot e^{i\phi(x,y)},
\end{align}

where $A(x,y)$ and $i\phi(x_{0},y_{0})$ are the amplitude and phase of the enhanced complex block. And $\phi(x_{0},y_{0})$ is taken as the final enhanced results.

The hardest part to transform these operations is that Gabor filters do not share weights on the whole image, but share on image blocks with same $\omega$ and $\theta$. To solve this problem, we propose a selective convolution method.

\paragraph{Grouped Phases}
Firstly, parameters are discretized into N different intervals and Gabor filters are generated respectively.
Then a group of filtered complex images can be obtained by convolving with these Gabor filters.
\begin{align}
    C(x,y,i)&=(I\ast g_{\omega_i,\theta_i})(x,y),~i=0,1,...,N-1
\end{align}
where $C(x,y,i)$ denotes the intensity value at pixel $(x,y)$ in the $i$-th filtered complex image. The grouped phases $F$ are the argument of group filtered images.
\begin{equation}
    F(x,y,i) = Arg[C(x,y,i)].
\end{equation}

\paragraph{Orientation Choose}
A mask is generated to select proper enhanced blocks from the grouped phases. The $i$th value at pixel $(x,y)$ in the mask $M$ is defined as,
\begin{align}
    M(x,y,i)&=
    \begin{cases}
        1,~if~\omega(x,y)=\omega_i,~\theta(x,y)=\theta_i \\
        0,~otherwise.
    \end{cases}
\end{align}

\paragraph{Enhanced Map}
Finally, the enhanced map can be calculated as,
\begin{align}
	E(x,y)&=\sum_{i=0}^{N}F(x,y,i) \cdot M(x,y,i)
\end{align}

This selective convolution can still be classified as a kind of ConvNet, since all operations are differentiable.

\subsubsection{Extraction}
Through template matching, minutiae can be easily extracted on phase~\cite{gao2010novel} or thinning images~\cite{maltoni2009handbook}. The minutiae score map $S$ can be computed as,
\begin{equation}
    S(x,y)=\max\limits_{t}(E\ast T_t)(x,y),
\end{equation}
where $T$ are the templates.
This module is equivalent to a ConvNet with one convolution layer and one maxout layer~\cite{goodfellow2013maxout}.

%\begin{figure}[t]
% \centering
% \includegraphics[width=0.8\linewidth]{ext_filter_bank}
% \caption{A group of extraction templates.}
% \label{fig:ext_filter_bank}
%\end{figure}

\subsection{Expand to FingerNet}
Plain FingerNet mentioned in Section~\ref{section:transform} can achieve a fair result on rolled/slap fingerprints, since it's typically designed for it.
However it failed on latent fingerprints.  It is not because the properties of fingerprints have changed when it comes to latent images, but the algorithms used to get those properties fail. This is caused by the contradiction between complex background noises and shallow ConvNet structures with poor expressive power.

%One natural idea is to use learnable deep ConvNets instead of handcrafted shallow ConvNets as the components of plain FingerNet.
Naturally, the simple network is then expanded with some convolutional layers to enhance its representation ability, and the weights are released to learn complex background variance from data.
Since the weights are initialized from simple network, the complete FingerNet won't perform worse.
%Inspired by recent works\cite{simonyan2014very,ioffe2015batch,chen2016deeplab} on ConvNets, we expand the FingerNet.

\begin{figure*}[t]
 \centering
 \includegraphics[width=0.9\linewidth]{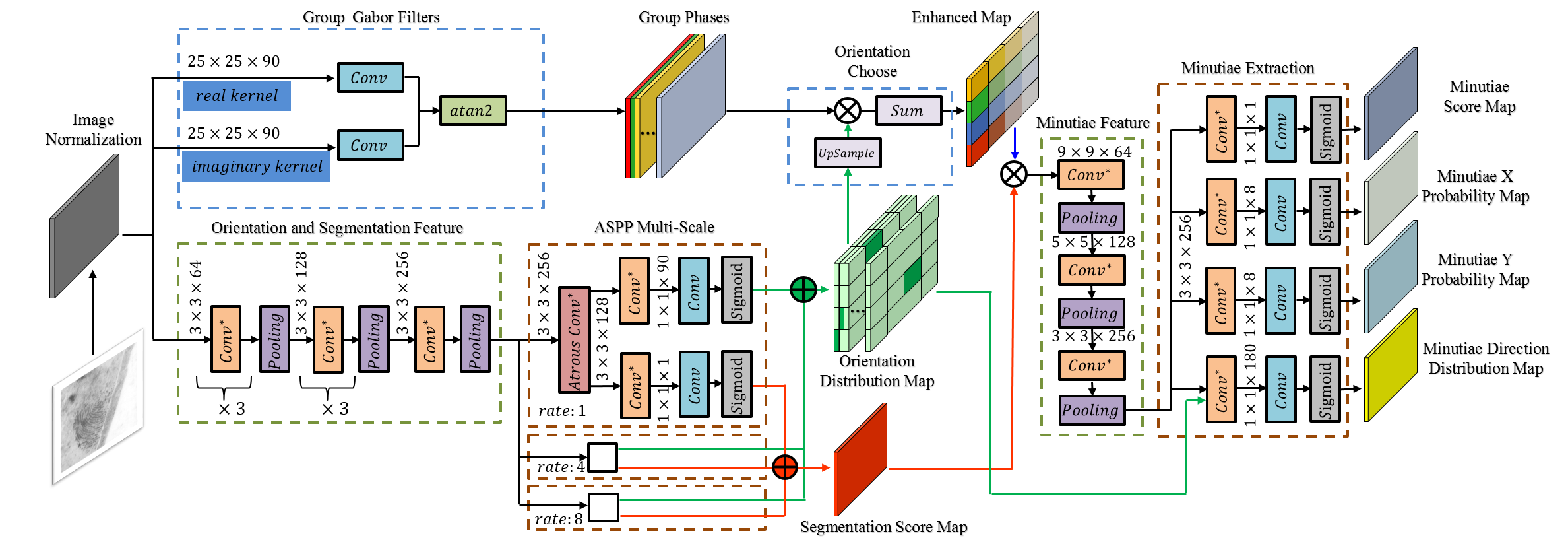}
 \caption{Our detailed FingerNet architecture. It is expanded from the plain FingerNet and able to be trained end-to-end.
 $Conv^*$ indicates a conv block containing convolutional, BatchNorm and PReLU layers.}
 \label{fig:detailed_pipeline}
\end{figure*}

The detailed architecture of FingerNet is shown in Fig~\ref{fig:detailed_pipeline}.
Next we discuss how to expand the plain network.
%FingerNet can be used universally on both good and bad quality fingerprints without any fine tuning.

\subsubsection{Normalization}
We directly adopt pixel-wise normalization mentioned in Section~\ref{section:fingerprint normalization} as our very beginning layer after image input.

\subsubsection{Orientation Estimation}
The deeper version of orientation estimation includes multi-scale feature extraction and orientation regression.

Basic feature extraction part has 3 conv-pooling blocks. Each conv-pooling block contains a pooling layer after a few convolutional blocks, while each conv block is made of a conv layer followed by a BatchNorm~\cite{ioffe2015batch} layer and a PReLU~\cite{he2015delving} layer.

Since multi-scale regularization~\cite{oliveira2008multiscale} will help, we adopt ASPP~\cite{chen2016deeplab} layer after basic feature extraction as our multi-scale solution. We use 3 atrous convolutional layers with different sample rates.

After that, a parallel orientation regression is carried on each scale feature maps and fused at last as the final estimation. Inspired by~\cite{gidaris2016locnet}, we let FingerNet directly predict the probabilities of $N$-discrete angles for each input pixel. The predicted angles at $(x,y)$ can be represented as a N-dimensional vector $p_{ori}=\{p_{ori}(i)\}_{i=0}^{N-1}$, where the $i$-th element $p_{ori}(i)$ indicates the probability of ridge orientation value of this position to be $\lfloor \frac{180}{N}\rfloor \cdot i$.

By doing so, we may get the final orientation output by either selecting a maximum response $\theta_{max}(x,y)$ or averaging to a more robust estimate~\cite{kass1987analyzing} $\theta_{ave}(x,y)$ as,
\begin{align}
\theta_{max}(x,y)&=\max_{i}{p_{ori}(i)}, \label{equation:theta_max}\\
\theta_{ave}(x,y)&=\frac{1}{2}atan2(\bar d_{sin}(x,y), \bar d_{cos}(x,y)), \label{equation:theta_rob}
\end{align}
where $\bar d(x,y)$ is the averaging ridge orientation vector and can be computed as,
\begin{align}
\bar d_{cos}(x,y)&=\frac{1}{N}\sum_{N}{p_{ori}(i) \cdot cos(2 \cdot \lfloor \frac{180}{N}\rfloor\cdot i)}, \\
\bar d_{sin}(x,y)&=\frac{1}{N}\sum_{N}{p_{ori}(i) \cdot sin(2 \cdot \lfloor \frac{180}{N}\rfloor\cdot i)}, \\
\bar d(x,y)&=\left[\bar d_{cos}(x,y), \bar d_{sin}(x,y) \right], \label{equation:d(x,y)}.
\end{align}

\subsubsection{Segmentation}
As mentioned in Section~\ref{section:segmentation}, learning based segmentation shares some features with orientation estimation. Hence for deeper version,  we directly let it share the entire multi-scale feature maps with orientation estimation part.

As for the classifier, we use a multi layer perception to predict the probability of each input pixel to be the region of interest, and output a segmentation score map with size of $\frac{H}{8} \times \frac{W}{8}$.

\subsubsection{Enhancement}
We directly adopt Gabor enhancement method mentioned in Section~\ref{section:gabor enhancement} as enhancement part for FingerNet. Considering ridge frequency in fingerprint is usually stable, we set ridge frequency to a fixed value and discretize ridge orientation to N intervals.

Different from plain FingerNet, the orientation distribution map is already an orientation mask. So the mask is multiplied directly by grouped phases. We just upsampled the orientation distribution map by the factor of 8 to fit the size of enhanced map.

%It is worth mentioning that the Gabor filters are initialized by standard Gabor function and set to be trainable.

\subsubsection{Minutiae Extraction}\label{section:minutiae extraction}
Minutiae extraction part takes enhancement output together with segmentation score map as input and conduct 3 conv-pooling blocks as feature extraction.
Then we generate 4 different maps for minutiae extraction.

The first map is minutiae score map, which represents the probability of each position $(x,y)$ to have a minutiae. Its size is $\frac{H}{8} \times \frac{W}{8}$.

The second and third maps are $X$ and $Y$ probability map. Since we only predict minutiae score map every 8 pixels, this position regression is essential to precise minutiae extraction. Inspired by~\cite{gidaris2016locnet}, we conduct a 8 disperse location prediction respectively for $X$ and $Y$ on each input feature point.

The fourth map is minutiae angle distribution map. It is completely the same as orientation distribution map, but the max angle value is changed from 180 to 360..

%In consideration of $X$ and $Y$ probability maps also provide minutiae location information, we conduct a fusion using this two maps and minutiae score map and take the fusion result as the final minutiae score map.

A minutiae list can be easily obtained by filtering minutiae score map with a proper threshold value. The precise location is acquired by adding the offset, which is the argument of the maximum $X$ and $Y$ probability.
The angle of minutiae is calculated using Eq.~\ref{equation:theta_max} or Eq.~\ref{equation:theta_rob}.

Since predicted minutiae may gather around, we use Non-maximum suppression(NMS) to clip redundant minutiae.

\subsection{Label, Loss and Training}

%After the traditional methods are transformed, the structure and initial weights are determined. Since we exactly know what each module is doing, a natural idea is to make each module doing its best. So we adopt a modular training method to make each module perform well on our database. And finally integrated them as a whole network to fine tuning.

\subsubsection{Weak, Strong and Ground Truth Label}

There is rare available labeled data of fingerprint orientation field or segmentation. Considering most latent databases are matched with rolled/slap fingerprints, we form 3 kinds of labels with different confidence.

Weak orientation labels are generated from matched rolled/slap fingerprints.
The matched pairs are from the same finger and share the same ridge structure.
The aligned rolled/slap fingerprints' orientation fields are fairly good estimations for corresponding latent fingerprints. %However, due to fingerprints' nonlinear distortion, it's not a hundred percent accurate.
We use minutiae to align fingerprint pairs and plain FingerNet to obtain rolled/slap fingerprints' orientation fields.

Weak segmentation labels are generated from minutiae convex hulls. The dilated and smoothed minutiae convex hulls are used as weak segmentation labels.

%Fig.~\ref{fig:weak_label} demonstrates an example for weak orientation label and segmentation label.

Since unoriented minutiae directions are the same as corresponding orientation fields, we take unoriented minutiae directions manually marked as our strong orientation labels.

Ground truth labels indicate 4 minutiae maps mentioned in Section~\ref{section:minutiae extraction}, which are transformed from manually marked minutiae list.

%It should be mentioned that labels for orientation distribution map and minutiae direction distribution map are not simply one-hot encoding angle values. The cross entropy loss between two one-hot vector is neither infinity or 0.
To measure the distance between angles and handle the discontinuity around $0^{\circ}$, we use inverted gaussian angle as label. The label $p_{\theta}=\{p_{\theta}(i)\}_{i=0}^{N-1}$ for $(x,y)$ position with angle $\theta$ can be computed as,
\begin{align}
p_{\theta}(i)&=p_{N(0,\sigma)}(min(|\theta-\lfloor\frac{\theta_{max}}{N}\rfloor\cdot i|,\notag \\
&\theta_{max}-|\theta-\lfloor\frac{\theta_{max}}{N}\rfloor\cdot i|)),i\in{[0,N)},
\end{align}
where $p_{N(0,\sigma)}(x)$ is the probability value of a gaussian distribution with mean 0 and variance $\sigma$ at $x$. $\theta_{max}$ is the max angle value, which is 180 for orientation and 360 for minutiae direction. With these kinds of labels, closer angles have smaller cross entropy loss and angles with same directional distance have same cross entropy loss.
%That is to say, the cross entropy loss between $0^{\circ}$ and $1^{\circ}$ is the same between $0^{\circ}$ and $179^{\circ}$.

%\begin{figure}[t]
%\centering
%\subfigure[rolled fingerprint]
%{\label{fig:1}\includegraphics[width=0.4\linewidth]{m22030_t}}
%\subfigure[latent fingerprint]{\label{fig:2}\includegraphics[width=0.4\linewidth]{m22030_l}}
%\subfigure[segmented orientation]{\label{fig:3}\includegraphics[width=0.4\linewidth]{o22030_t_gt}}
%\subfigure[segmented orientation]
%{\label{fig:4}\includegraphics[width=0.4\linewidth]{o22030_l_gt}}
%\subfigure[segmentation]{\label{fig:3}\includegraphics[width=0.4\linewidth]{seg22030}}
%\subfigure[calculated segmented orientation]{\label{fig:3}\includegraphics[width=0.4\linewidth]{o22030_l_cal}}
%\caption{Flowchart of weak label preparation}
%\label{fig:weak_label}
%\end{figure}

\begin{figure}[t]
 \centering
 \includegraphics[width=1\linewidth]{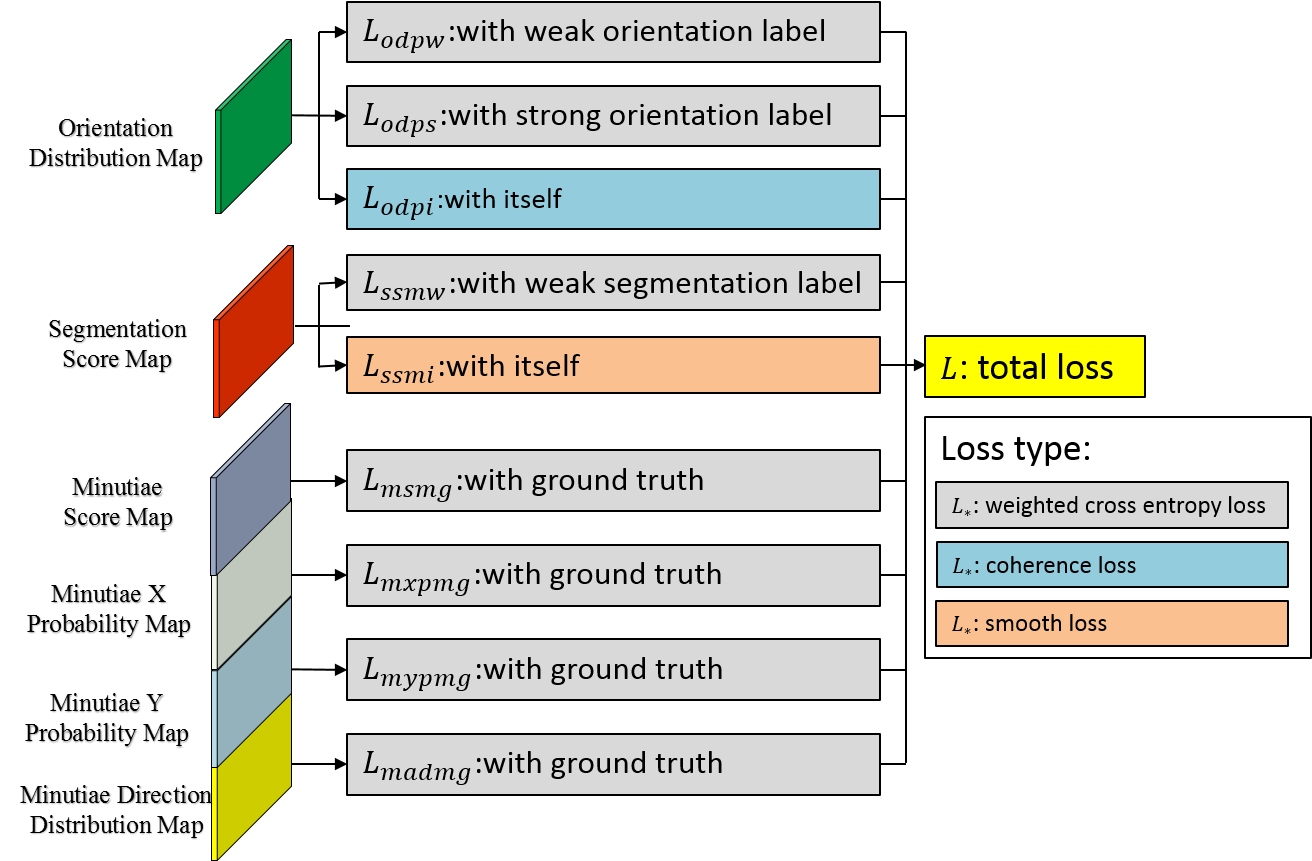}
 \caption{The loss cluster illustrating our total loss. It is a weighted sum of 9 different losses coming from orientation, segmentation and minutiae extraction.}
 \label{fig:lossfunction}
\end{figure}

\subsubsection{Loss Definition and Training Procedure}

The loss cluster is shown in Fig.~\ref{fig:lossfunction}. The total loss $L$ is a weighted sum of 9 different losses.
%and can be computed as,
%\begin{align}
%L&=\lambda_{1} \cdot L_{odpw} + \lambda_{2} \cdot L_{odps} + \lambda_{3} \cdot L_{odpi} \notag \\
% &+ \lambda_{4} \cdot L_{ssmw} + \lambda_{5} \cdot L_{ssmi} + \lambda_{6} \cdot L_{msmg} \notag \\
% &+ \lambda_{7} \cdot L_{mxpmg} + \lambda_{8} \cdot L_{mypmg} + \lambda_{9} \cdot L_{madmg}
%\end{align}

As shown in Fig.~\ref{fig:lossfunction}, there are only 3 different types of loss. The cross entropy loss is defined as,
\begin{align}
L_{*}&=-\frac{1}{|ROI|}\sum_{ROI}\sum_{i=1}^{N}(\lambda^{+} p_{l_*}(x,y) log(p_{*}(i|(x,y))) \notag \\
	 &+\lambda^{-} (1-p_{l_*}(x,y)) log(1-p_{*}(i|(x,y)))),
\end{align}
where ROI is the region of interest, $\lambda^{+}$ and $\lambda^{-}$ are weights for positive and negative samples, $p_{l_*}(x,y)$ and $p_{*}(i|(x,y))$ are the probability values at $(x,y)$ in label map and predicted map respectively.
%Since some labels are not defined out of ROI, like orientation on background and minutiae direction on none minutiae area, we only calculate losses in ROI.
Since positive and negative labels are always unbalanced, we use $\lambda^{+}$ and $\lambda^{-}$ to balance their loss contributions.

Orientation coherence~\cite{kass1987analyzing} is a strong domain prior knowledge, so we turn it into a loss function to constrain the orientation distribution map. It can be calculated as,
\begin{align}
\bar d &= \{\bar d(x,y)\}_{x,y},~|\bar d| = \{|\bar d(x,y)|\}_{x,y}, \notag \\
Coh &= \frac{\bar d \ast J_3}{|\bar d| \ast J_3}, \notag \\
L_{odpi}&= \frac{|ROI|}{\sum_{ROI}Coh} - 1,
\end{align}
where $J_3$ is an all-ones matrix with size of $3 \times 3$ and $\bar d$ is the orientation vector mentioned in Eq.\ref{equation:d(x,y)}.

In order to make segmentation more smooth with less noises and outliers, we simply try to suppress the edge responses. It can be calculated as,
\begin{align}
L_{ssmi}&=\frac{1}{|I|}\sum_{I}|M_{ss} \ast K_{lap}|,
\end{align}
where $M_{ss}$ is segmentation score map, $K_{lap}$ is a laplace edge detection kernel and $I$ is the region of total image.

After model construction and data preparation, we conduct a two step training procedure. Firstly, let FingerNet learn ridge properties by training with orientation and segmentation losses. After a few epoches, we add minutiae losses. The idea is to let FingerNet learn step by step. Adam optimizer is adopted and other detailed parameter settings can be found in our open source FingerNet codes.

%-------------------------------------------------------------------------
\section{Experiments}

We compare minutiae extraction performance with other algorithms on different quality fingerprint databases to test FingerNet's generalization ability.
As can be seen from the following experiments, our unified FingerNet can calculate reliable orientation field, segmentation, enhanced fingerprint and minutiae without any fine tuning operation.

\subsection{Database}

The training data was collected from crime scenes, including about 8000 pairs of matched rolled fingerprints and latent fingerprints. Each latent fingerprint is $512\times{512}$ pixels in size and 500 pixels per inch (ppi) with expert marked minutiae. FingerNet is trained on this database and remains the same in the following experiments.

Our test experiments are conducted on NIST SD27~\cite{garris2000nist} and FVC 2004 database set A~\cite{maio2004fvc2004}. NIST SD27 contains 258 latent fingerprints with minutiae marked by experts. Each fingerprint is  $768\times{800}$ pixels in size and 500 ppi.
FVC 2004 database contains 3600 rolled fingerprints. These fingerprints are also 500 ppi but different in image size.
%Since our net is implemented as a fully convolutional manner~\cite{long2015fully}, it can satisfy different size of input images.

\begin{figure}[t]
 \centering
 \includegraphics[width=0.85\linewidth]{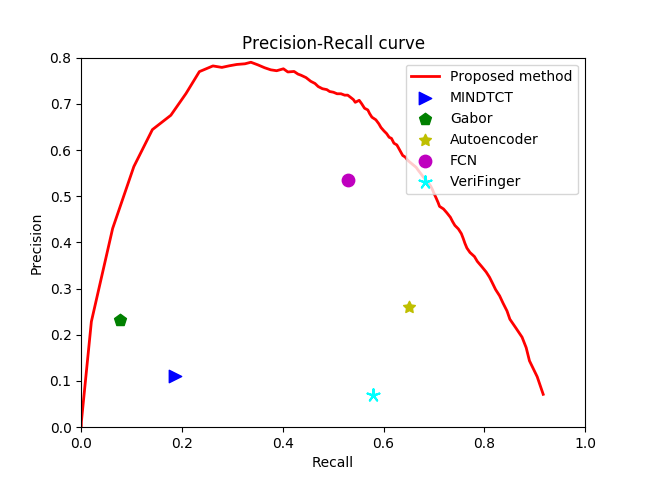}
 \caption{Precision-Recall curves of different minutiae extraction algorithms on NIST SD27.}
 \label{fig:PRC}
\end{figure}

\subsection{Minutiae Extraction Performance}

The performance of minutiae extraction is evaluated with Precision-Recall curve. Precision is defined as positive predictive value and recall is defined as true positive rate.
An extracted minutia is assigned to be true if its distance to a manually labeled minutia is less than 15 pixels, and the angle between the two is less than $30^{\circ}$. Furthermore, this is one to one match.

Fig.~\ref{fig:PRC} compares the minutiae extraction performance with other methods on NIST SD27.
MINDTCT is an open source minutiae extractor from NIST Biometric Image Software~\cite{watson2007user}.
Gabor-based algorithm~\cite{gao2010novel} extracts minutiae on Gabor phase.
AutoEncoder-based algorithm~\cite{sankaran2014latent} extracts minutiae with a learned stacked denoising sparse autoencoder.
%The input fingerprints are manually segmented and their orientations are not calculated.
FCN-based algorithm~\cite{tang2016latent} extract minutiae with a learned fully convolutional network.
VeriFinger~\cite{verifingerneuro} is a well-known commercial system used for minutiae extraction and fingerprint matching.

The mean error of location and angle are 4.4 pixels and 5.0$^{\circ}$ respectively. For segmentation results compared with weak labels, the true positive rate is 0.88 and true negative rate is 0.92. For orientation results compared with weak labels, the accuracy is 0.87 within $20^{\circ}$.
The recall can't reach to 1 due to segmentation and non-maximum suppression.
About 0.6 seconds is used on average to extract minutiae on NIST SD27.

\begin{figure}[t]
 \centering
 \includegraphics[width=0.85\linewidth]{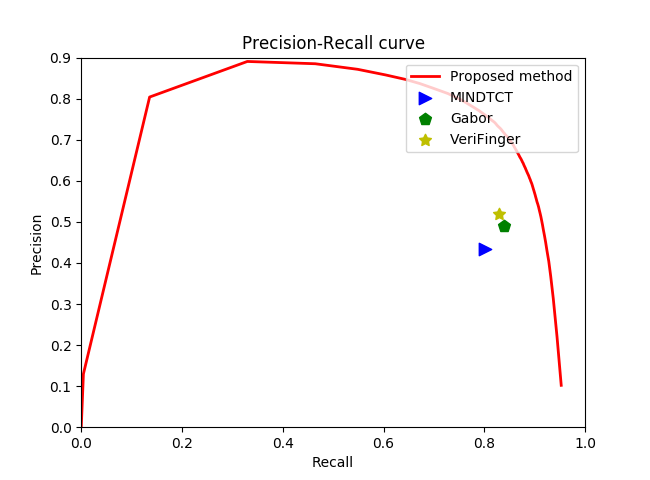}
 \caption{Precision-Recall curves of different minutiae extraction algorithms on FVC 2004 database set A.}
 \label{fig:PRC_FVC}
\end{figure}

Fig.~\ref{fig:PRC_FVC} compares the minutiae extraction performance on FVC 2004 database. The mean error of location and angle are 3.4 pixels and 6.4$^{\circ}$ respectively.

\begin{figure*}[t]
\centering
\subfigure[fingerprints]{\label{fig:1}\includegraphics[width=0.19\linewidth]{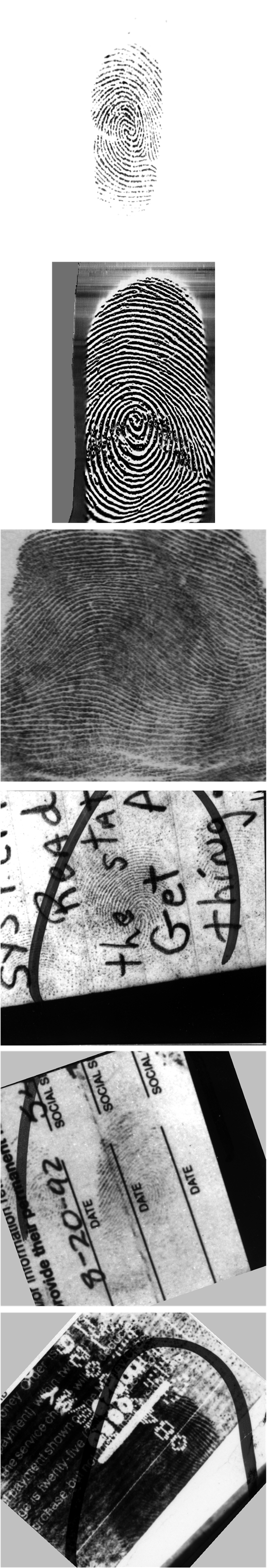}}
\subfigure[segmented orientation field]{\label{fig:2}\includegraphics[width=0.19\linewidth]{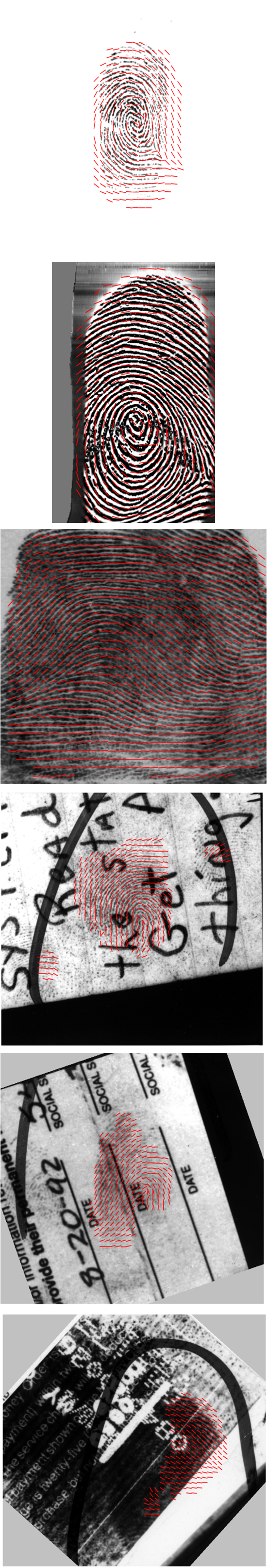}}
\subfigure[enhanced fingerprints]{\label{fig:3}\includegraphics[width=0.19\linewidth]{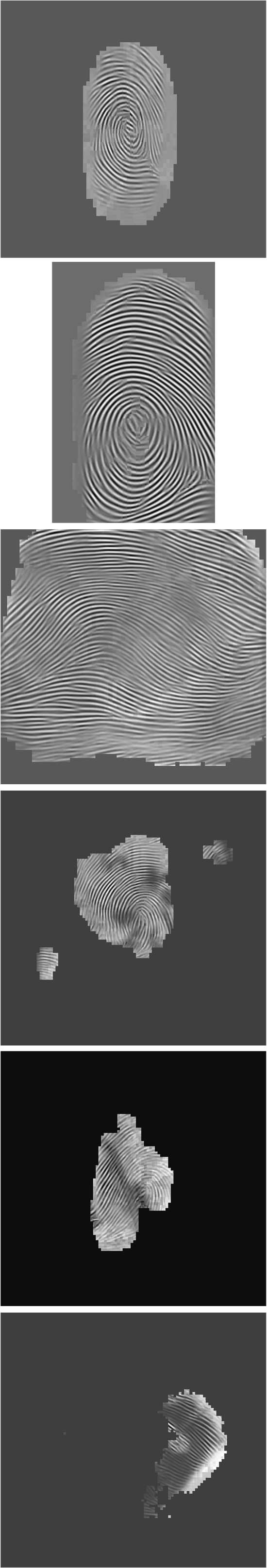}}
\subfigure[mintuiae map]{\label{fig:4}\includegraphics[width=0.19\linewidth]{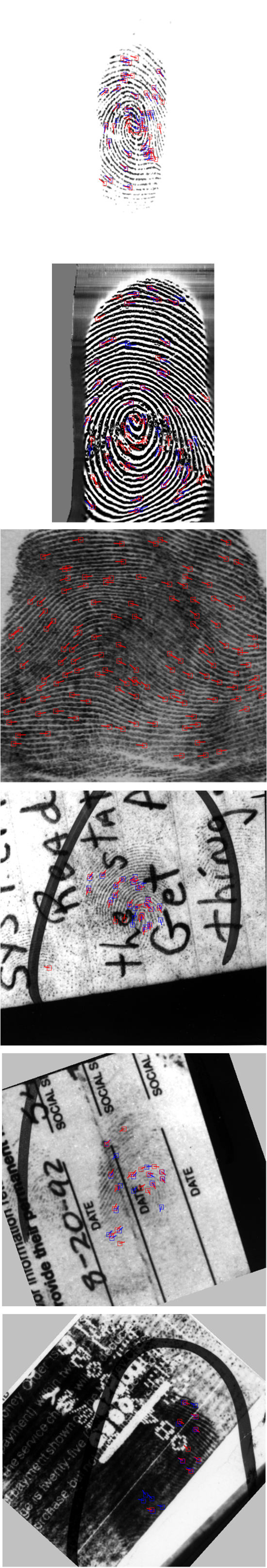}}
\caption{More results of our proposed FingerNet. Column (a)-(d) are original fingerprints, segmented orientation field, enhanced fingerprints and minutiae map. From top to bottom, fingerprints are sampled from FVC 2004 DB1A, FVC 2004 DB3A, NIST 4, NIT SD27 with good, bad and ugly quality. In (d), red squares denote our extracted minutiae and blue squares denote manually marked minutiae.}
\label{fig:sample_result}
\end{figure*}

Fig.~\ref{fig:sample_result} shows our performance on several fingerprints with different quality. They are sampled from FVC 2004 database, NIST 4 database~\cite{watson1992nist} and NIST SD27.

\begin{figure}[t]
 \centering
 \includegraphics[width=0.9\linewidth]{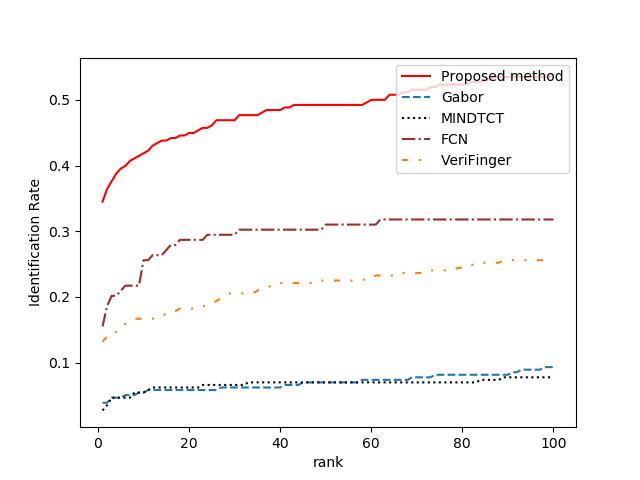}
 \caption{Identification performance (CMC curves) of different minutiae extraction algorithms on NIST SD27.}
 \label{fig:identification}
\end{figure}

\subsection{Identification Performance}
Fig.~\ref{fig:identification} shows Cumulative Match Characteristic curves on NIST SD27 to test whether fingerprint matching can benefit from FingerNet.
%Especially in bad quality fingerprints.
The matching algorithm is based on extended clique models~\cite{fu2013extended}. Only minutiae are used in this method.
The gallery contains about 40K fingerprints including NIST SD27, NIST SD4 and our in-house database.
Result shows that FingerNet outperforms other methods.

\section{Conclusion and Future Work}

%Minutiae extraction on latent fingerprints is of critical importance in criminal investigation. Previous works on rolled/slap fingerprints failed on latent fingerprints due to noisy ridge pattern and presence of background noise. Some learning based methods haven't took advantages of domain prior knowledge and the performance is limited.
We propose a new way to guide the deep network's structure design and weight initialization for combining domain knowledge and deep learning representation ability. We demonstrate the pipeline consisting of several typical traditional methods is equivalent to a simple network with fixed weights.
The network is then expanded and the weights are released while preserving end-to-end differentiability.
Following this idea, FingerNet is proposed for efficient and reliable minutiae extraction on both rolled/slap and latent fingerprints. This algorithm has combined domain knowledge and deep learning method to outperform other minutiae extraction algorithms.

Future work will include (1) integrating ridge frequency to the pipeline, (2) exploring more accurate segmentation algorithm, and (3) extending FingerNet to matching.

\section*{Acknowledgment}
This work was supported by NSFC(61333015).

\end{document}